\documentclass[runningheads]{llncs}

\usepackage{eccv}

\usepackage{eccvabbrv}

\usepackage{graphicx}
\usepackage{booktabs}

\usepackage[accsupp]{axessibility}  % Improves PDF readability for those with disabilities.

\usepackage[pagebackref,breaklinks,colorlinks]{hyperref}
\usepackage{orcidlink}
\usepackage{multirow}

\begin{document}

\title{WSI-VQA: Interpreting Whole Slide Images by Generative Visual Question Answering} 

\titlerunning{WSI-VQA}

\author{Pingyi Chen\inst{1,2,3}\orcidlink{0000-0001-5569-5725}\and
Chenglu Zhu\inst{2,3} \orcidlink{0000-0001-5705-3718}\and
Sunyi Zheng\inst{2,3} \orcidlink{0000-0002-9005-4875} \and Honglin Li \inst{1,2,3} 
\and Lin Yang\inst{2,3}}

\authorrunning{P. Chen et al.}

\institute{Zhejiang University \and
Research Center for Industries of the Future, Westlake University\and School of Engineering, Westlake University.
\\
\email{\{chenpingyi,yanglin\}@westlake.edu.cn}}

\maketitle

\begin{abstract}
 Whole slide imaging is routinely adopted for carcinoma diagnosis and prognosis. Abundant experience is required for pathologists to achieve accurate and reliable diagnostic results of whole slide images (WSI). The huge size and heterogeneous features of WSIs make the workflow of pathological reading extremely time-consuming. In this paper, we propose a novel framework (WSI-VQA) to interpret WSIs by generative visual question answering. WSI-VQA shows universality by reframing various kinds of slide-level tasks in a question-answering pattern, in which pathologists can achieve immunohistochemical grading, survival prediction, and tumor subtyping following human-machine interaction. Furthermore, we establish a WSI-VQA dataset which contains 8672 slide-level question-answering pairs with 977 WSIs. Besides the ability to deal with different slide-level tasks, our generative model which is named Wsi2Text Transformer (W2T) outperforms existing discriminative models in medical correctness, which reveals the potential of our model to be applied in the clinical scenario. Additionally, we also visualize the co-attention mapping between word embeddings and WSIs as an intuitive explanation for diagnostic results. The dataset and related code are available at \href{https://github.com/cpystan/WSI-VQA}{https://github.com/cpystan/WSI-VQA}.
  \keywords{Whole Slide imaging \and Visual Question Answering}
\end{abstract}

\section{Introduction}
\label{sec:intro}

Multimodal large language models (MLLM), such as GPT-4V \cite{gpt4v}, LLaVa \cite{llava}, Qwen-VL \cite{qwenlv}, have demonstrated remarkable superiority over a wide range of visual-language tasks. Even in medical domains requiring specialized knowledge and where a large visual gap exists, MLLMs continue to exhibit exceptional performance \cite{gpt4v,pmcclip,llavamed,clip,biomedclip}.  Despite their impressive performance in the medical area, these MLLMs can not be easily applied to achieve visual understanding for WSIs which are characterized by a tremendous size (the maximum can be 100,000×100,000 pixels at 20$\times$ magnification). Recent years have also witnessed the emergence of pathological MLLMs (MI-Zero \cite{lu2023mizero}, PLIP \cite{huang2023plip}, CONCH \cite{CONCH}). Although a large number of pathological images are seen by the MLLMs, the patch-level pre-training makes it challenging to achieve promising performance on slide-level tasks which requires an understanding of the whole WSI like carcinoma subtyping.

In VQA for natural images, the input with the approximate size of $256\times256$ is worth $16\times16$ words with visually descriptive labels. In the pathology field, the slide-level prediction for WSIs is a knotty problem that is usually formulated in a multiple-instance learning (MIL) framework \cite{shao2021transmil,zhang2022dtfd,ilse2018attention,chikontwe2020multiple,li2021dual}. Patches are cropped from the whole image followed by reducing dimensions to make computation feasible. Patch embeddings act as individual instances which are then aggregated to generate the final result in a weakly-supervised manner. The whole image acts as a bag exhibiting enormous heterogeneity with labels that are visually hard to distinguish. Differentiating tumors, as a basic slide-level task, depends on the fine-grained visual features that localize in a very small region. Survival outcome prediction is a challenging regression task where the field of view needs to be expanded to the entire WSI and to understand the interactions of different components (stroma, lymphocyte, microenvironment, etc).

Visual-language MLLMs unify a variety of visual tasks such as object segmentation, scene understanding, and optical character recognition by elaborately building appropriate prompts and obtaining textual descriptions \cite{gpt4v,llava,qwenlv}. In the pathology field, previous task-specialized MIL models can only solve certain problems like tumor grading despite their remarkable performance. To promote the development of visual language learning for whole slide images, we propose a novel paradigm of WSI-VQA in which the input is the whole slide image and the output is the text. In WSI-VQA, a visual encoder is applied to extract and aggregate patch-level embeddings, and a text encoder gets word embeddings of the questions. Patch embeddings and word embeddings are aligned by a co-attention mapping in the decoder to generate the answers. In contrast to those VQA approaches \cite{vqa_chen,vqa_liu,vqa_lu,vqa_over,vqa_reduce} which introduce a classifier to select the right answer given a finite set of choices, our generative model can predict the answer word by word in free form, which enables our model to deal with diverse problems especially when contrastive or classification candidate answers are not available in clinical practice.

In the field of pathology, visual-language models lag behind due to insufficient training data for MLLMs. The large resolution and visually challenging features make it especially difficult to annotate WSIs. In addition, the concern about data privacy and security also prevents enough slide-level data from being widely accessible. We are glad to see that some researchers collect pathological text from a wide range of resources like books \cite{micaption} and social media (Twitter) \cite{huang2023plip}. However, these textual descriptions are constrained to narrating pathological patches instead of the whole slide image, which makes models struggle to achieve promising slide-level performance like tumor subtyping \cite{lu2023mizero}.

To effectively train our proposed WSI-VQA models, we propose an automatic and scalable pipeline to curate a slide-level question-answering dataset, termed WSI-VQA. It contains 977 WSIs equipped with 8672 question-answering pairs. Our VQA pairs are generated based on well-established WSI captions \cite{wsicaption} and TCGA clinical information (TCGA is the largest publicly available database for cancers), covering a variety of WSI-related tasks. It brings several benefits. Firstly, since the labels are well-annotated by experts in TCGA, the pathological text is reliable and provides abundant slide-level information. Secondly, the correspondence between WSIs and the text is clear and well-aligned. Last but not least, considering related data is already de-identified and public, we have no problem opening our collected dataset. In our experiment, we train and evaluate our generative model on the collected WSI-VQA as the benchmark. According to the results, our approach demonstrates superiority by unifying different clinical tasks and surpassing previous multiple instance learning (MIL) methods which are specially designed for WSI classification. 
To the best of our knowledge, our collected dataset is the first slide-level question-answering dataset and our approach is also the pioneering work of WSI-VQA. Our contributions can be summarized as follows:

\begin{enumerate}    
    \item We propose a novel framework of WSI-VQA which reframes various slide-level tasks in the pathology field into the question-answering paradigm. Our generative model can achieve competitive performance on these different WSI-related tasks. The universality and scalability of our approach reveal its potential as the foundation MLLM for computational pathology.
    
    \item We propose the WSI-VQA dataset, which is named WSI-VQA. It contains 977 WSIs and their corresponding QA pairs. The dataset which is going to be publicly available will promote the development of pathological visual-language learning.

    \item Visualizing the co-attention map improves the interpretability of our model and can provide clinical clues for pathologists.

\end{enumerate}

\section{Dataset Construction}
In this section, we will provide more details about the construction process of our proposed dataset WSI-VQA, as shown in Fig. \ref{fig:framework}. We will first give a brief introduction to the source data and comprehensively explain the scalable generation process of question-answering pairs, followed by analyzing and visualizing data statistics.

\subsection{Data Generation}
\begin{figure}[tb]
  \centering
  \includegraphics[width=\linewidth]{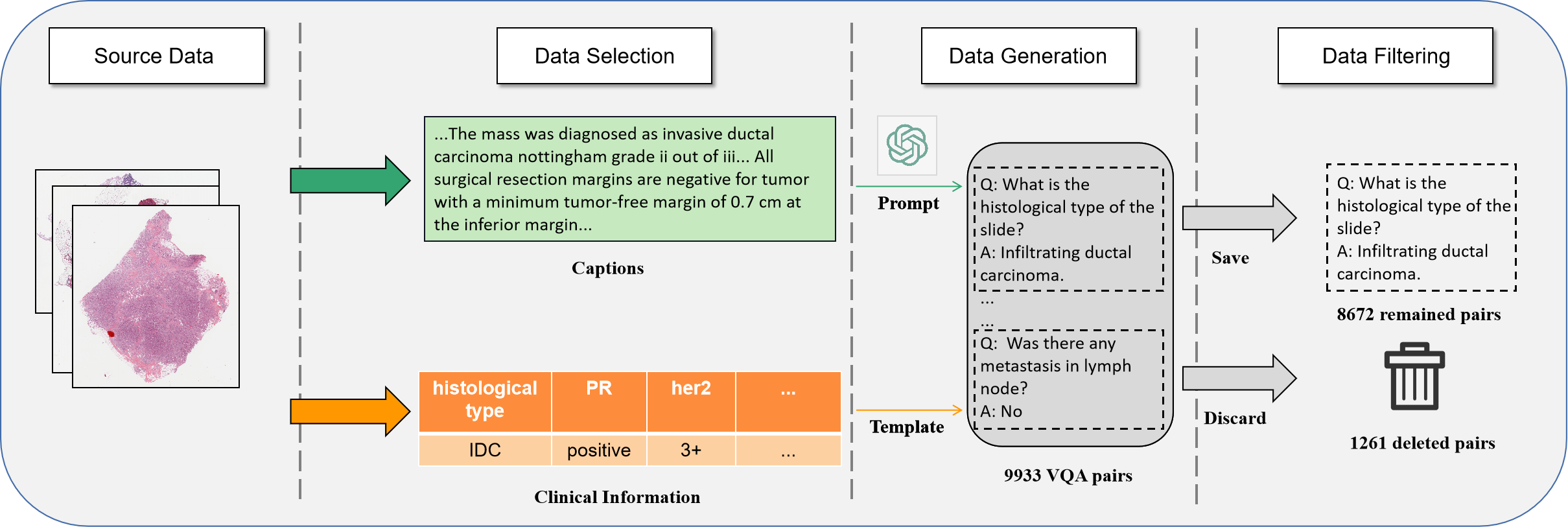}
  \caption{Flow diagram of the WSI-VQA construction. We first select pathological captions and clinical indexes in TCGA. Then, close-ended and open-ended VQA pairs are obtained through GPT and fixing templates respectively. Finally, clinical validation is adopted to remove the pairs which are flawed or can not be inferred from the WSI.
  }
  \label{fig:framework}
\end{figure}

TCGA establishes an infrastructure which collects and characterizes a large number of cancer samples. It contains diagnostic WSIs and relevant clinical information of various modalities. We resort to TCGA-BRCA (the largest subset of TCGA about breast cancer) to curate our slide-level VQA dataset because of its reliability and security. From the captions of TCGA, as shown in Fig. \ref{fig:framework}, we generate question-answering pairs equipped with multiple choices with the aid of LLMs. The other question-answering pairs are obtained from the clinical files in TCGA by extracting keywords to fit templates.

\subsubsection{Close-ended Subset.} Based on the largest WSI caption dataset which summarized pathological reports from TCGA \cite{micaption}, we can automatically generate several questions accompanied with candidate answers for each caption with the aid of large language models (LLM) like GPT\cite{gpt3}. Following the work \cite{pmcvqa}, we use the following prompt to generate WSI-VQA pairs:

\textit{Ask 6 questions about the content and generate four options for each question. The output should use the following template: i:‘the question index’ question:‘the generate question’ choice: ‘A: option content B: option content C: option content D: option content’ answer: The correct option.}

Although this approach makes the automation of VQA pair generation feasible, the problem of hallucination brought by LLMs can not be ignored. The hallucination means that LLMs may generate content that is contradictory to the common knowledge or unfaithful to the provided content. Therefore, we adopt data filtering to discard those pairs that own significant defects. In addition, the answers that can not be inferred from WSIs are also removed from our collected dataset. Finally, we obtained 4535 close-ended VQA pairs which not only contain questions and answers but also provide candidate choices. Since the close-ended subset has choices for each question, we can compute the medical accuracy to measure the performance of our generative model.

\subsubsection{Open-ended Subset.} TCGA-BRCA includes various clinical indexes: subtyping, immunohistochemical testing, and survival prediction. By matching the keywords and their corresponding values in the question-answering templates, we can obtain high-quality VQA pairs which contain abundant clinical information. We designed several question-answering templates and one of the templates is shown below:

\textbf{Q: What is the result of [KEY]? A: [VALUE].}

During the training process, we randomly select a template for each pair. To make our generative model reasonable, we discard those clinical indexes that can not be inferred from diagnostic WSIs. As a result, the open-ended subset contains 4137 VQA pairs.

\begin{figure}[tb]
  \centering
  \includegraphics[width=\linewidth]{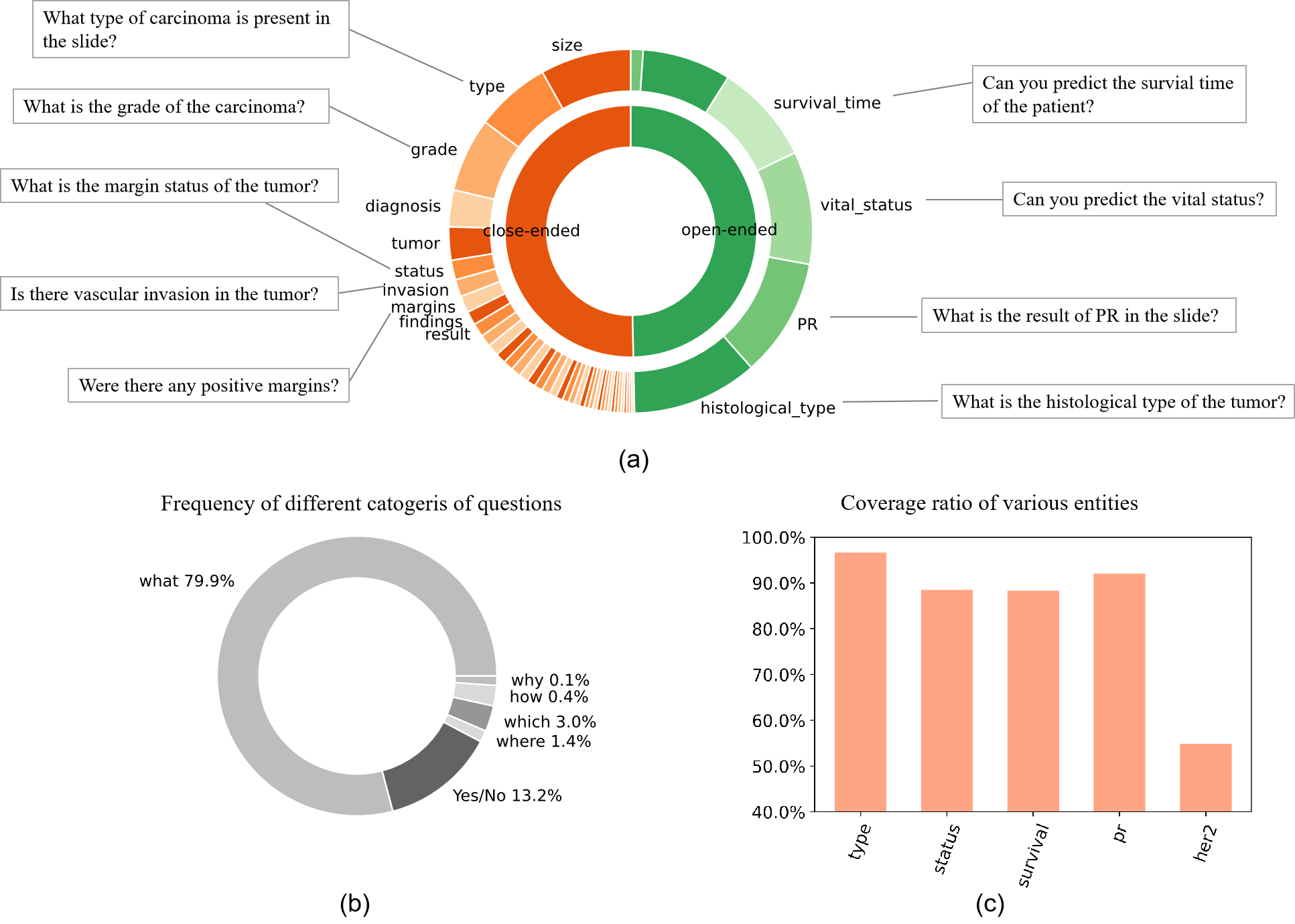}
  \caption{(a) The distribution of biological entities in our proposed dataset. The center circle shows the distribution of close-ended pairs and open-ended pairs.  The outer circle presents the entities in the dataset from which we can see that our questions cover a wide range of slide-level characteristics. Several examples of questions are also demonstrated. (b) The frequency of different categories of questions set by the first word. The 'what' questions dominate the VQA pairs with a frequency of $79.9\%$. (c) The ratio of various entities in the open-ended subset. 'Her-2' and 'PR' are two biomarkers for breast cancer.
  }
  \label{fig:dis}
\end{figure}

\subsubsection{Data Statistics.}
We provide a statistical overview of our proposed dataset. The dataset contains 977 WSIs and 8672 QA pairs where each WSI is associated with an average of 8.9 QA pairs. Among these QA pairs, the close-ended subset has 4535 QA pairs while the open-ended one has 4137 pairs. As shown in Fig. \ref{fig:dis} (a), the questions in the dataset demonstrate a diversity ranging from the questions that require spatial information such as identifying the margin status to challenging questions like immunohistochemical grading of the WSI. Fig. \ref{fig:dis} (b) shows the distribution of different categories of questions. The dataset is predominantly characterized by 'what' questions, accounting for $80\%$ of its content. There are also other types of questions such as $13.2\%$ 'Yes/No' questions, $1.4\%$ 'where' questions, and $3\%$ 'which' questions. Since the open-ended subset is collected from the clinical files in TCGA, we calculate the coverage ratio of different entities in Fig. \ref{fig:dis} (c). For example, the coverage ratio of 'her2' is around $50\%$, which means nearly half of the cases in the open-end subset are not provided with Her-2 results.

\section{Method}
\begin{figure}[tb]
  \centering
  \includegraphics[width=\linewidth]{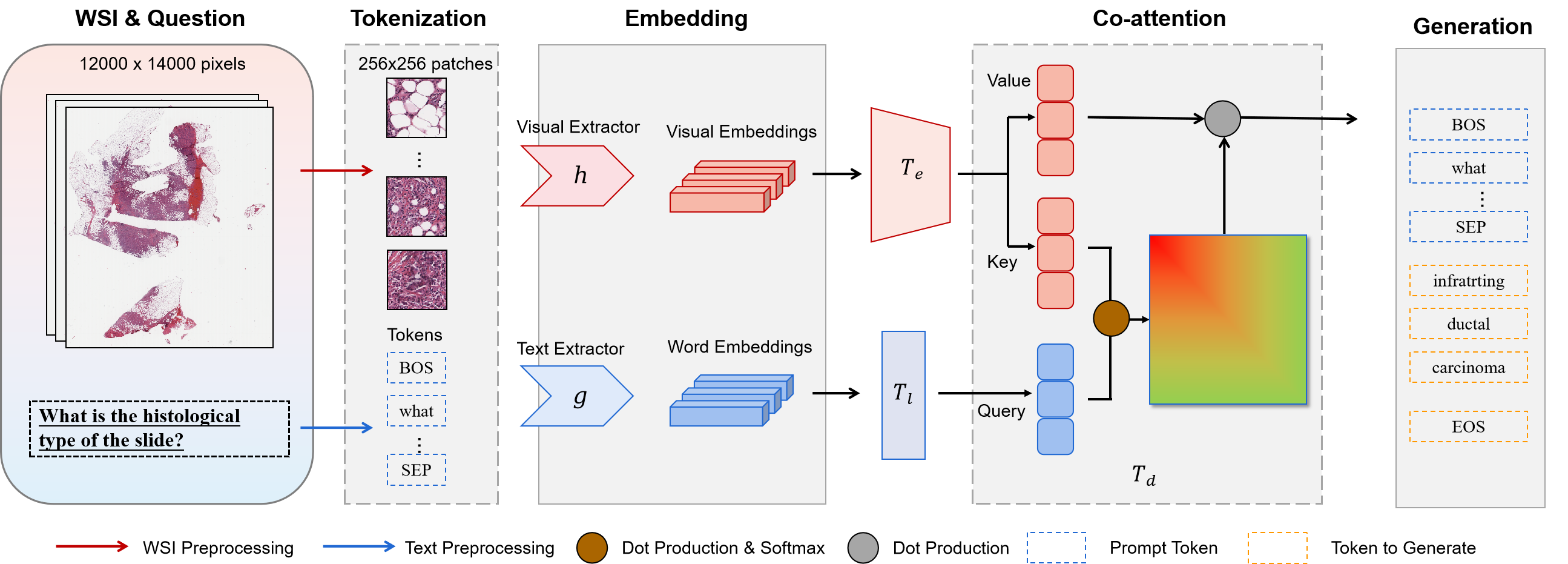}
  \caption{Sturcture of our proposed VQA model. First, the WSI and the question are processed and tokenized for subsequent stages. Then, various visual and text extractors are applied to extract embeddings given visual and word tokens. $T_e$ stands for the transformer encoder and $T_l$ represents the linear mapping, aiming to achieve alignment between visual and word embeddings. The interaction mechanism between different modalities, which is instantiated as the co-attention in the transformer decoder $T_d$, is adopted to generate the answer.
  }
  \label{fig:model}
\end{figure}

In this section, we elaborate on our WSI-VQA model, the Wsi2Text Transformer (W2T) the structure of which is visualized in Fig. \ref{fig:model}. Firstly, we explain the theoretical base of our VQA method in Section \ref{theory_base}.  We introduce our way of pre-processing for WSIs and the formulation of patch embeddings and word embeddings in Section \ref{preprocessing}. In Section \ref{interaction}, we present the interaction between visual and textual features showing how word embeddings attend to relevant patches when responding to a question. 
\subsection{Problem Formulation}
\label{theory_base}
The inputs $(X_i, Q_i, Y_i)$ are the whole slide image $X_i$ and the question $Q_i$, predicting $Y_i$ as the output.
The large resolution necessitates the pre-processing of WSIs.  The WSI can be treated as the aggregation of a group of instances which are patches with a much smaller resolution. It is denoted as $X_i = \{x_i^j\}_{j=1}^{M_i}$ where $X_i$ is the $i-th$ WSI  and $M_i$ is the sequence length which is usually larger than 10000.  A visual extractor $h$ is adopted to extract patch embeddings, denoted as $h(\{x_i^j\})\in \mathbb{R}^{M_i\times l}$ where $l$ is the embedding size.

The question can be seen as a sequence of tokens after tokenization, which is formulated as $Q_i=\{q_i^j\}_{j=1}^{T_i}$ where $T_i$ is the number of tokens. We use a text extractor $g$ to transform the tokens into word embeddings, denoted as $g(\{q_i^j\})\in \mathbb{R}^{T_i\times k}$ where $k$ is the embedding size for word embeddings.

A generative model is incorporated to generate the answer $Y_i=\{y_i^j\}_{j=1}^{N_i}$ where $N_i$ is the length of target sequence. The model parameterized by $\phi$ is trained with language model loss objectives that seek the maximization of the sum of conditional possibilities of words in the sequence. The negative log-likelihood (NLL) loss can be formulated as:

\begin{equation}
    L= -\sum_i\sum_{t=1}^N log(y_t|h(\{x_i^j\}),g(\{q_i^j\}),\{y_i^j\}_{j<t};\phi).
\label{equation:loss}
\end{equation}

Given the patch embeddings, word embeddings, and the previous sequence $\{y_i^j\}_{j<t}$, the probability of the next word to be generated can be calculated.

\subsection{Preprocessing for Bag Construction}
\label{preprocessing}

\subsubsection{Visual Embeddings.}

In natural VQA, it is GPU-feasible to directly feed the image so that the visual embeddings are trainable.  In the pathology field, we follow the curriculum of previous MIL approaches where the large image is cut down into small individual patches. We first perform preprocessing to remove the non-informative regions which make up 70\% of a typical whole slide image. In specific,  the whole slide image is transformed in the color space from  RGB to HSV at low objectives. Then, we use thresholding to filter out the white regions based on the saturation channel. Then, we crop the non-overlapped patches on the foreground region with a sliding window.  

After obtaining the patches  $ \{x_i^j\}_{j=1}^{M_i}$, we apply pre-trained neural networks to extract features from these patches.  The $256\times256$ patches are transformed to $512$-dim visual embeddings using ResNet, largely reducing the complexity of the computation. It is noteworthy that the visual extractor is non-trainable.

\subsubsection{Word Embeddings.}
The sequence of the question is much shorter than visual embeddings. In the cohort of our questions, the average sequence length is 9.2 tokens, with some questions having up to 24 tokens. Therefore, it is feasible to train the word embeddings from scratch. In our experiments, we also use two well-established language models as the text extractor: BioClinicalBert \cite{pubmedbert} and PubMedBert \cite{bioclinicalbert} which are trained with abundant medical corpus, proving to yield excellent word embeddings.

\subsection{Interaction between WSI and Text}
\label{interaction}
Current multimodal approaches in the pathology field devote efforts to linking mutually independent patches with corresponding textual descriptions \cite{huang2023plip,lu2023mizero}, which limits their ability on slide-level tasks since the WSI is the aggregation of a lot of mutually relevant patches and their collected descriptions are focused on patch-level narration. In our framework, by formulating the WSI and the text as bag representations  $E_X$ and $E_Q$,  we can not only model the interaction among patches cropped from the same WSI but also establish fine-grained interactions between patch tokens and word tokens. After obtaining the visual embeddings  $E_X\in \mathbb{R}^{M_i\times l}$, a transformer encoder is adopted to capture the mutual instance relationship, which is formulated as below:

\begin{equation}
    SelfAtt(X) = softmax(\frac{QK^T}{\sqrt{d}
    })V
    =softmax(\frac{W_Q E_X E_X^T W_K^T}{\sqrt{d}})W_VE_X,
\end{equation}

where $W_Q, W_K, W_V\in \mathbb{R}^{l\times l}$ are trainable parameters and $d$ is a scaling factor. We denote the transformer’s \cite{vaswani2017transformer} query, key, and value as $Q$, $K$, and $V$ respectively. In the transformer encoder $T_e$, our model learns to allocate attention to the instances and builds the mutual-instance correlations for the subsequent decoding stage. It is also a mimic of clinical workflow where pathologists usually observe by switching the region of interest and yield the diagnosis based on both cellular-level features and the structural information of tissues and organs.

The co-attention mechanism in the standard transformer decoder is adopted to capture the relations between visual and text concepts, which can be formulated as:

\begin{equation}
\begin{split}
    CoAtt(X,Q) = softmax(\frac{QK^T}{\sqrt{d}
    })V
    =&softmax(\frac{W_Q E_Q E_X^T W_K^T}{\sqrt{d}})W_VE_X  \\= \textbf{A}W_VE_X,
\end{split}
\end{equation}

where $W_Q, W_K, W_V\in \mathbb{R}^{l\times l}$ are trainable matrices multiplied by the visual and word embeddings. The word embeddings need to be aligned to transform their size from $k$ to $l$ by the linear mapping $T_l$ before being fed to the transformer decoder. The co-attention module $\textbf{A} \in \mathbb{R}^{T_i\times M_i}$ learns fine-grained patch-word similarity showing how much word tokens $\{q_i\}$ attend to patches $\{x_i\}$. For a single word embedding $q_n$, the according weight matrice is $\textbf{A}_n \in \mathbb{R}^{1\times M_i} $ which reflects how attentions are allocated to the enormous patches based on the $n-th$ word token. For each WSI-question pair, we choose the keyword (biological entity in the question) to guide the generation of attention maps as a visual explanation for the answer. For example, since the keyword 'her2' is the $4-th$ token in the question as shown in Fig. \ref{fig:att}, we choose $\textbf{A}_4\in \mathbb{R}^{1\times M_i} $ to be visualized as heatmaps. The highlighted regions prove to be more relevant to the answer.

\section{Experiments}
In this section, we first provide more details about our proposed dataset WSI-VQA. The backbones we choose to extract visual and word embeddings are also introduced. Finally, we elaborate on the setting of our proposed model W2T.

\subsection{WSI-VQA Dataset}
Our proposed WSI-VQA dataset contains a total of 977 WSIs with 8671 question-answering pairs. We split the dataset into a training set, a validation set, and a test set. The training set includes 804 WSIs with 7139 question-answering pairs. The validation set contains 87 WSIs and 798 pairs. And the test set contains 86 WSIs and 735 question-answering pairs. The right choices in the close-ended subset when testing include 151 for A, 107 for B, 86 for C, and 46 for D. We use both language metrics and clinical metrics to evaluate our model. For the closed-ended subset, we also adopt accuracy (ACC) by comparing the sentence similarity to measure the performance of the model.

\subsection{Visual Extractors}
Here we present different visual extractors in our generative model.
\textbf{ResNet} \cite{resnet}
 is a classic vision model, pre-trained by a lot of natural images and their classification label in ImageNet. The version we use is ResNet-50 which is composed of 50 convolutional layers with 25.6M parameters. \textbf{ViT} \cite{vit} is the vision transformer VIT-S which has 12 transformer layers with approximately 22M parameters.
\textbf{DINO} \cite{dino} is a strategy of self-supervised pre-training with a standard ViT model. It is adopted to pre-train the backbone with our cropped patches from TCGA-BRCA. We choose VIT-S as the backbone.
\textbf{HIPT} \cite{hipt} which conducts hierarchical self-supervised learning with a pyramid transformer is a WSI-specialized self-supervised approach of pre-training. It exploits the hierarchical structure inherent in WSIs from the whole TCGA to learn high-resolution image representations.

\subsection{Text Extractors}
In this section, we present different text extractors which are employed in W2T.
\textbf{BioClinicalBert} \cite{bioclinicalbert} is a BERT-like model which is trained on all notes from MIMIC III\cite{mimiciii}, a database (text size 0.5B words / 3.7GB) which has abundant electronic health records from the hospitals. It is initialized with BERT-Base which has 12 transformer layers with around 110M parameters.
\textbf{PubMedBert} \cite{pubmedbert} is trained with abstracts and articles from PubMed and PubMedCentral \cite{pubmedcentral} with the size of 3.1B words / 21G based on BERT-base structure. 
\textbf{Scratch} means that we use an embedding mapping to extract the text features without freezing the parameters. Since the length of word tokens is much smaller than image tokens, it is feasible to train a text extractor from scratch while the visual extractor can not be trainable.

\subsection{Training Setting}
The transformer encoder $T_e$ and decoder $T_d$ include three layers and the hidden size is 512. The multi-head self-attention has 4 heads and the embedding size is also 512. Adam is adopted as the optimizer to train our model with a learning rate of 1e-4 and a weight decay of 5e-5. In the inference stage, the size of beam search is 3. All our experiments are trained on one A100-80G Nvidia GPU with PyTorch.

\section{Results}

\subsection{Benchmark}
As illustrated in Table \ref{tab:comp}, we introduce our WSI-VQA benchmark on our proposed dataset. Different visual extractors and text extractors are included for comparison. To measure the performance of the baselines, we consider two kinds of metrics: 1) natural language generation (NLG) metrics such as BLEU \cite{papineni2002bleu}, METEOR \cite{banerjee2005meteor}, and ROUGE \cite{lin-2004-rouge} which are widely used in natural language processing tasks, and 2) factual consistency and completeness (FCC) metrics to measure the clinical performance of the model. It is worth noting that ACC is only adopted to evaluate the close-ended subset which contains multiple candidate choices one of which is the right answer. And $Fact_{ent}$ \cite{miura2020improving} is used to measure how much the generated entities in the pathology domain are consistent with the reference.

\begin{table*}[t]
\caption{Quantitative results of the VQA performance of different baselines in WSI-VQA. Different combinations of visual extractors and text extractors are present for comparison. BLUE-n indicates the BLUE score computed based on n-grams. NLG metrics also contain METEOR and ROUGE. ACC and $Fact_{ent}$ are adopted to measure the factual consistency and completeness of the answer.}
\renewcommand\arraystretch{1.3}
  \centering
  
  \resizebox{\linewidth}{!}
{
    \scriptsize
  \begin{tabular}{l|l|cccc|cc}
   \toprule
        \multirow{2}{*}{Visual Extractor}& \multirow{2}{*}{Text Extractor}&\multicolumn{4}{c}{NLG Metrics} & \multicolumn{2}{c}{FCC Metrics}\\
       
     &&Bleu-1&Bleu-4&METEOR&ROUGE &ACC&$Fact_{ent}$\\
   \midrule
 \multirow{3}{*}{ResNet}&PubMedBert& 36.9&19.1&22.0&43.2&43.1&\textbf{92.0}\\
&BioClinicalBert&34.1 & \textbf{20.6} & 21.6 & 42.9& \textbf{46.9}&91.2\\
&Scratch&
\textbf{38.3} & 19.7 & \textbf{23.2} & \textbf{46.2}& 43.3&91.1\\
\hline

 \multirow{3}{*}{ViT}&PubMedBert&32.3 &14.7&19.8&42.0&48.8&92.0\\
&BioClinicalBert&30.2 & 13.1 & 19.9 & 41.0& \textbf{48.5} &\textbf{92.9}\\
&Scratch&\textbf{35.9}&\textbf{21.3} & \textbf{22.5}& \textbf{45.6}&45.4 &90.8\\
\hline

 \multirow{3}{*}{DINO}&PubMedBert& 35.4&17.7&19.6&41.7&50.4&\textbf{91.5}\\
&BioClinicalBert& 34.4&15.9 &21.1 &43.2 & 49.9 &90.9\\
&Scratch&\textbf{38.6} & \textbf{18.3} & \textbf{22.7} & \textbf{44.8}& \textbf{50.8}&91.0\\
\hline

 \multirow{3}{*}{HIPT}&PubMedBert&33.0 & 18.1&19.9&41.8&49.7&91.5\\
&BioClinicalBert&31.6 & 14.3 & 20.2 & 40.0& 48.4 &\textbf{94.1}\\
&Scratch&\textbf{34.2}& \textbf{19.6} & \textbf{21.1} & \textbf{44.0}& \textbf{50.1}&92.1\\

\bottomrule
  \end{tabular}}

\label{tab:comp}
\end{table*}

We find that the best result for NLG Metrics is achieved when the text extractor is training from scratch. In specific, the increases of $+2.4$ Bleu-1, $+0.6$ Bleu-4, $+1.2$ METEOR, and $+3.0$ ROUGE are observed on WSI-VQA when compared against the model with PubMedBert given visual embeddings from ResNet. Training the text extractor from scratch shows its superiority over generating fluent and readable answers while those task-agnostic word embeddings lag in terms of the NLG metrics. It may be because of the domain gap between the corpus for pre-training and the pathological corpus that is used in our experiment. However, we find it interesting that the 'scratch' strategy does not win consistently regards the FCC metrics. For example, an increase of $+4.1$ ACC and $+2.1$ $Fact_{ent}$ when BioClinicalBert is adopted given the visual embeddings from ViT is observed, which reveals the potential of those well-established text extractors in recognizing and understanding the clinical concepts.  

In terms of the visual extractor, we observe that DINO and HIPT (in-domain pre-training) consistently beat ResNet and ViT (out-of-domain pre-training) in medical correctness. In specific, a large increase of $+2.4$, $+1.4$, and $+5.4$ ACC are present respectively when DINO pre-training extractor is adopted compared to ViT which is pre-trained with ImageNet. It is reasonable that the visual extractors pre-trained with in-domain data are more capable of extracting relevant features in the pathological patches, which contributes much to the diagnosis of WSIs. For all the baselines, the ACC metrics are nearly 50, and the $Fact_{ent}$ metrics are over 90, which shows the potential of WSI-VQA in clinical scenarios.

\begin{figure}[tb]
  \centering
  \includegraphics[width=\linewidth]{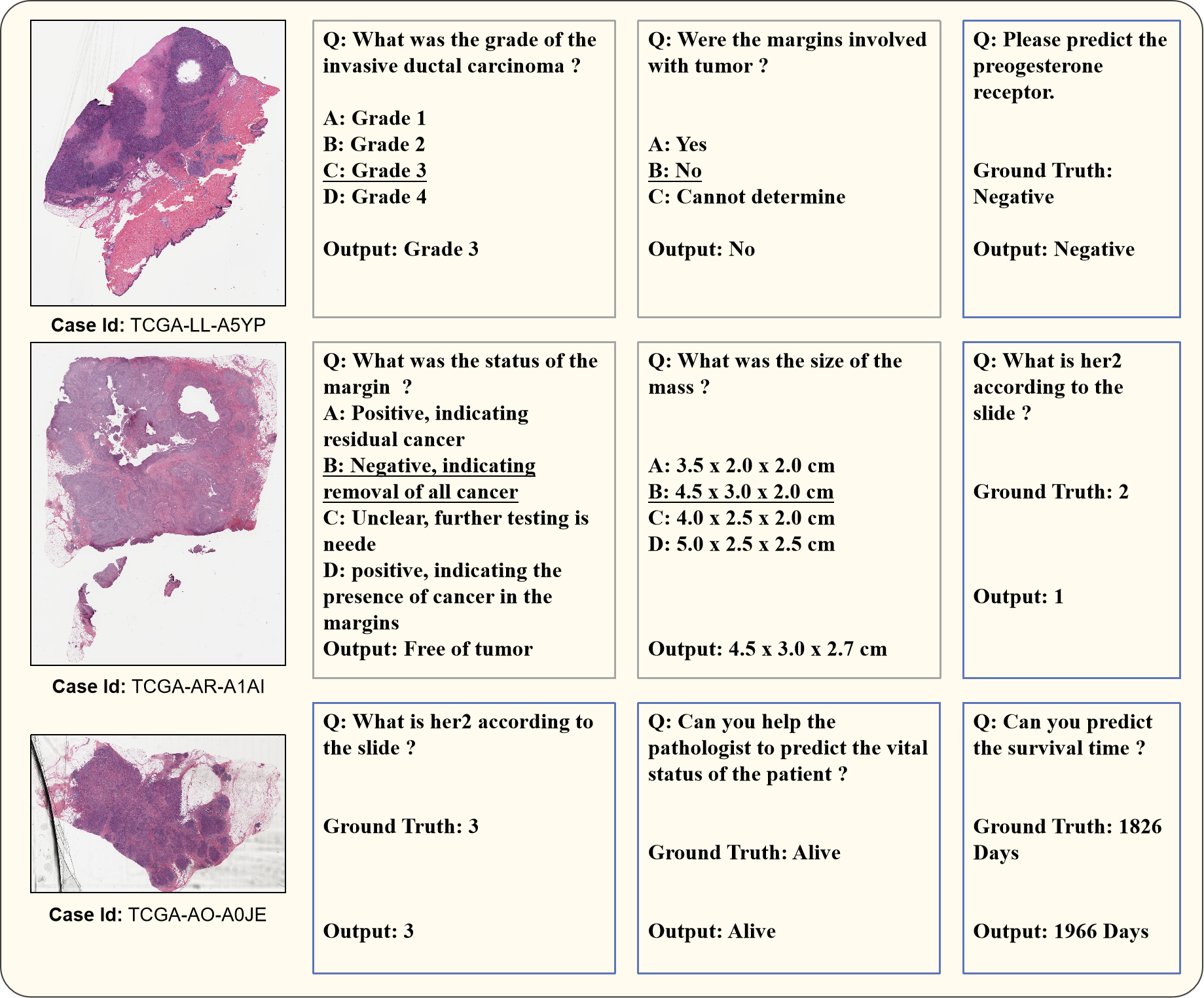}
  \caption{Several examples of the WSIs and their corresponding VQA pairs. The pairs in the grey rectangle are from the close-ended subset which has multiple choices while the ones in the blue rectangle are from the open-ended subset. The choice which is underlined is the right answer. These questions are all challenging because they require sufficient medical knowledge and understanding of complex characteristics in the gigapixel WSIs.
  }
  \label{fig:qa}
\end{figure}

\begin{table*}[t]
\caption{Quantitative results of different slide-level tasks: histological subtyping, PR prediction, and survival prediction. For the binary classification tasks, we adopt precision (P), recall (R), and F1 score as the evaluation metrics. For the regression task, we choose the c-index which is widely adopted to measure the performance of survival prediction.}
\renewcommand\arraystretch{1.3}

  \resizebox{0.9\linewidth}{!}
{
    \scriptsize
      \centering
  \begin{tabular}{l|ccc|ccc|c}
   \toprule
        \multirow{2}{*}{Method}& \multicolumn{3}{c}{Subtyping}& \multicolumn{3}{c}{PR}& Survival\\
        \cline{2-8}
        &P&R&F1 &P&R&F1&C-index \\

   \midrule
    Max-pooling &59.8&70.1&64.4&81.8&84.9&83.3&54.7\\
    AB-MIL&74.8 &81.5 &77.1&86.6&84.1&85.2&52.5\\
    CLAM-SB & 74.2&84.1&79.7&86.5&84.9&85.7&55.4\\
    Trans-MIL & 78.0&83.6&80.6&82.3&79.7&80.6&56.1\\
    \hline
    W2T (ResNet+scratch) &81.9 &85.1&83.3&\textbf{85.1}&82.2&84.0&56.3\\
    W2T (DINO+scratch) & \textbf{82.3}&\textbf{87.8}&\textbf{85.2}&84.3&\textbf{87.8}&\textbf{86.5}&\textbf{59.7}\\

\bottomrule
  \end{tabular}}

\label{tab:clinical}
\end{table*}

\begin{figure}[tb]
  \centering
  \includegraphics[width=\linewidth]{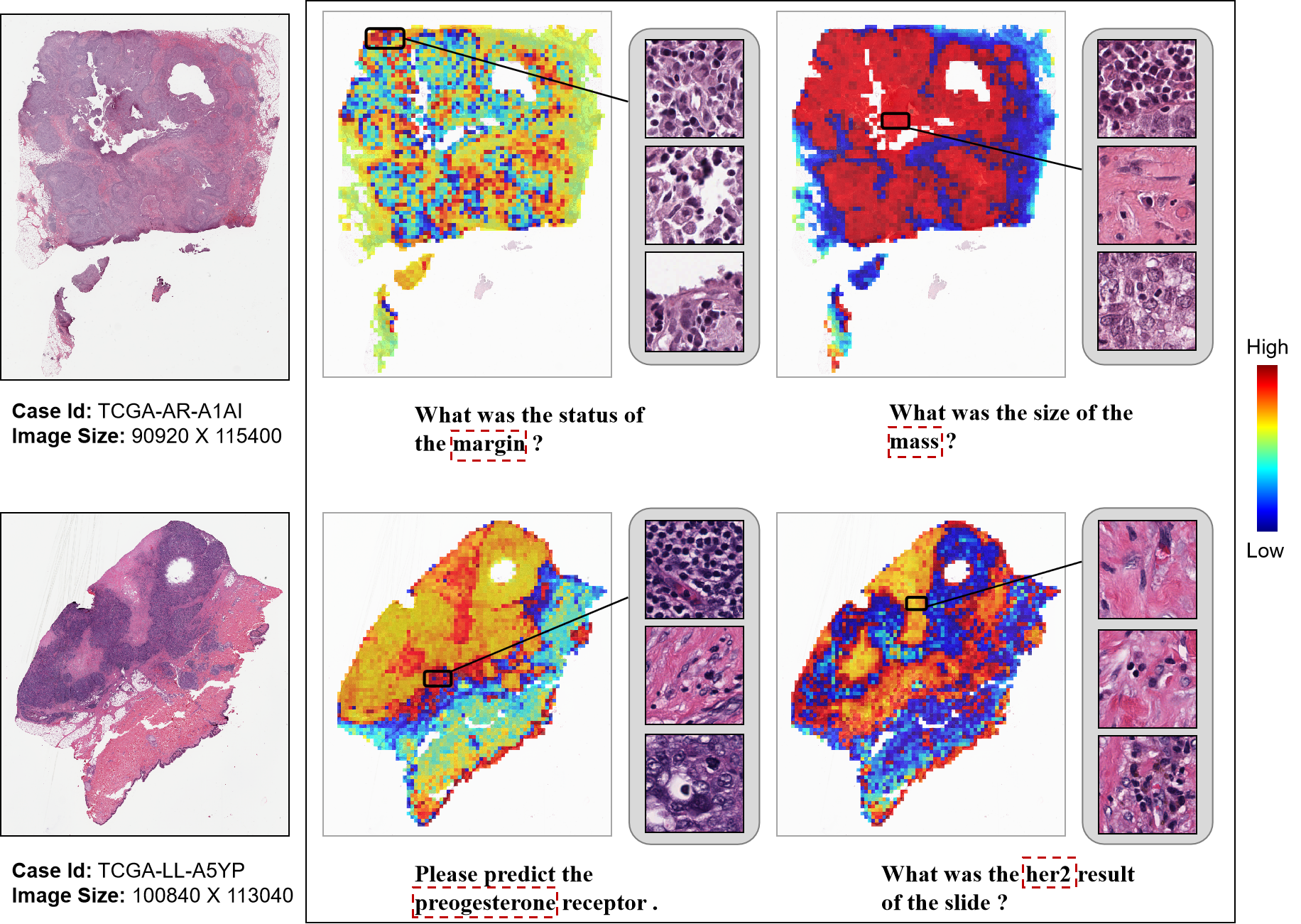}
  \caption{Co-attention visualization with corresponding questions. The keyword in the question, which is marked in a red rectangle, guides the interaction and aggregation of image patches using co-attention weights. The heatmap reflects how much each patch in the WSI is relevant to the word token ranging from blue to red.
  }
  \label{fig:att}
\end{figure}

\subsection{Slide-level Tasks}

Recent years have witnessed the development of AI-aided digital pathology and some deep-learning models can even achieve human-level performance in terms of slide-level tasks\cite{zhang2019pathologist}.  As a multimodal WSI-VQA method, our method can deal with various kinds of slide-level tasks if given appropriate prompts. Therefore, we evaluated the performance of our model on certain clinical tasks and re-produced several multiple instance learning (MIL) methods which are widely adopted in WSI classification for comparison, as shown in Table \ref{tab:clinical}. The MIL approaches which are elaborately designed for processing WSIs include: AB-MIL \cite{chikontwe2020multiple}, CLAM-SB \cite{lu2021clam}, and TransMIL \cite{shao2021transmil}. 

In terms of the histological subtyping, our proposed framework (W2T) surpasses all the MIL approaches. We observe an increase of $+3.9$ precision, $+1.0$ recall, and $+2.7$ F1 score of our ResNet-equipped model compared to the second-best MIL method. If the visual extractor is pre-trained with DINO, a more advanced performance can be seen where the F1 score is $85.2$. Progesterone receptor (PR) is the hormone receptor and serves as the basis for endocrine therapy. In terms of the prediction of PR, our model still achieves competitive performance compared to MIL approaches. Survival prediction is challenging in computational pathology since it is an ordinal regression task and involves capturing complex features in WSIs. Our proposed approach with ResNet or DINO as the visual extractor obtains the c-index of $56.3$ and $59.7$ respectively. 

We also present several examples which are shown in Fig. \ref{fig:qa}. The questions are all challenging slide-level tasks that require specialized pathological knowledge and a deep understanding of the features of the WSI. For the question of the margin status, it is essential to localize the related region and recognize the fine-grained features like tissue morphology and cell distribution. Whereas, grading the carcinoma needs to look through the whole image and capture the structural information.  It is worth noting that our model can handle various tasks by only adjusting the prompts and can also achieve promising performance, which reveals the great potential of our framework as the foundation model in computational pathology.

\subsection{Visualization}
The co-attention heatmap which reflects the interaction between visual concepts and word embeddings can act as an intuitive explanation for the decision-making procedure and provide clinical clues for pathologists. We overlay the co-attention weights which attend to the keyword in the question with the thumbnail of the corresponding WSI. The image patches that have high attention weights are highlighted in the heatmap with red color.

As illustrated in Fig. \ref{fig:att}, we choose two cases with four co-attention maps. We observe that our generative model can attend to the regions of interest according to the questions. When being asked about the status of the margin, it focuses on the margin of the tumor. If given the prompt about the mass size, our model expands its view to the whole tumor in the slide, which reflects the spatial awareness of our model. For the immunohistochemical (IHC) tasks, our model allocates more attention to the tumor-associated stroma than the regions which show high-grade tumor morphology. For the PR prediction, regions which have high attention are focused on the tumor cells and micro-environment like tumor stroma, which nevertheless are ignored when predicting the Her-2 result.  In the clinical scenario, IHC results such as Her-2 can not be directly obtained from H$\&$E images. However, many recent works \cite{her2_conde2022herohe,her2_farahmand,her2_slidegraph+}, which achieved the prediction given only the WSI, have proved the connection between H$\&$E stained images and IHC results. Since the connection needs further investigation,  I believe our method can provide relevant clinical clues and make AI approaches more interpretable to pathologists.

\section{Limitations and Future Work}
As a generative framework, our model may suffer from hallucination which is frequently observed in large language models especially when we scale up our model. The black-box property of AI-aided methods and the possibility of generating unreasonable answers may be the major obstacles to being applied in the clinical. Although we have managed to improve the interpretability like visualizing the co-attention map, there is still a long way to go before we earn the trust of pathologists. In addition, since the current framework relies on the pre-trained visual extractor, such task-agnostic embeddings may hinder the performance of our model. In the future, we are going to scale up our pipeline to build a more powerful and general MLLM for computational pathology.

\section{Conclusion}
In this paper, we propose interpreting gigapixel whole slide images by generative visual question answering (WSI-VQA). Our approach demonstrates significant superiority in managing a broad range of slide-level tasks and achieving remarkable performance on these tasks, which reveals the potential of our method as the foundation model for computation pathology due to its scalability. The pathologists can know what they want to know or have difficulty with by giving appropriate prompts to the model. The explanation of the decision-making can be visualized in the format of heatmaps which makes our model more transparent and convincing. On the data end, we curate the first WSI-VQA dataset which is going to be public. The dataset will promote the development of MLLMs in the pathology field. In addition, other fields that involve large-resolution modalities can also be inspired by our work.

\textbf{Acknowledgement.}This study was partially supported by the National Natural Science Foundation of China (Grant no. 92270108), Zhejiang Provincial Natural Science Foundation of China (Grant no. XHD23F0201), and the Research Center for Industries of the Future (RCIF) at Westlake University.

\bibliographystyle{splncs04}
\bibliography{egbib}
\end{document}